\documentclass[11pt]{article}

\usepackage{PRIMEarxiv}
\usepackage{titlesec}
\usepackage{longtable}
\usepackage{caption}
\usepackage{float}
\usepackage[utf8]{inputenc} 
\usepackage[T1]{fontenc}    
\usepackage{hyperref}       
\usepackage{url}            
\usepackage{booktabs}       
\usepackage{amsfonts}       
\usepackage{lmodern,textcomp}
\usepackage{eurosym}
\usepackage{amsmath, amssymb}
\usepackage{nicefrac}       
\usepackage{microtype}      
\usepackage{lipsum}
\usepackage{fancyhdr}       
\usepackage{graphicx}       
\usepackage{appendix}
\usepackage{rotating}
\usepackage{adjustbox}
\usepackage{multirow}
\graphicspath{{media/}}     
\titlespacing\section{0pt}{12pt plus 4pt minus 2pt}{0pt plus 2pt minus 2pt}

\title{Synthesis of separation processes with reinforcement learning}

\author{
  Stephan C.P.A. van Kalmthout \\
  University of Twente \\
  Enschede, The Netherlands \\
  \texttt{s.van.kalmt@gmail.com} \\
 \\
   \And
  Laurence I. Midgley \\
  Instadeep \\
  London, United Kingdom \\
  \texttt{laurencemidgley@gmail.com} \\
  \And
  Meik B. Franke \\
  University of Twente \\
  Enschede, The Netherlands \\
  \texttt{m.b.franke@utwente.nl} \\
}

\begin{document}
\maketitle

\begin{abstract}
This paper shows the implementation of reinforcement learning (RL) in commercial flowsheet simulator software (Aspen Plus V12) for designing and optimising a distillation sequence.\footnote{Code and thesis are available at https://github.com/lollcat/Aspen-RL} The aim of the SAC agent was to separate a hydrocarbon mixture in its individual components by utilising distillation. While doing so it tries to maximise the profit produced by the distillation sequence. All actions of the agent were set by the SAC agent in Python and communicated in Aspen Plus via an API. Here the distillation column was simulated by use of the build-in RADFRAC column. With this a connection was established for data transfer between Python and Aspen and the agent succeeded to show learning behaviour, while increasing profit. Although results were generated, the use of Aspen was slow (190 hours) and Aspen was found unsuitable for parallelisation. This makes that Aspen is incompatible for solving RL problems.
\end{abstract}


\section{Introduction}
Reinforcement learning is a sub-field of machine learning, in which an agent aims to learn an optimal policy which maximises its expected reward. The policy is improved by setting actions inside the environment and updating the agent based on these experiences. Previously RL has shown to be capable of outperforming humans in Chess and Go \cite{SilverASelf-playb}, but also shown success in process control \cite{Hoskins1992ProcessLearning, Shin2019ReinforcementControl}. Recent research has shown the applicability of RL in process synthesis  \cite{Midgley2019ReinforcementSynthesis,Midgley2020DeepSynthesisb,Stops2022FlowsheetNetworksb,Gottl2021AutomatedConcept,Gottl2022AutomatedLearning}. Reinforcement learning shows potential in handling open-end problems. Where superstructures have their optimal solution embedded in the user-defined superstructure, allows the reinforcement learning for exploration outside this pre-defined superstructure.

\section{Reinforcement learning}
\label{sec:headings}

\begin{figure}[!ht]
    \centering
    \includegraphics[scale=0.55]{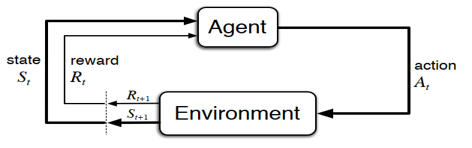}
    \caption{Markov Decision Process schematic \cite{Sutton2018ReinforcementIntroduction}}
    \label{MDP}
\end{figure}

A key characteristic of an RL problem is that it is defined as a Markov Decision Process (MDP). This implies that the choice of the current action only depends on the current state, see figure \ref{MDP}. An MDP consists of 2 entities, an agent and environment. The agent observes the state of the environment, \(S_{t}\), and sets an action accordingly, \(A_t\). This action is selected from the possible actions defined by policy \(\pi\). Due to the action, the environment transitions to its new state, \(s_{t+1}\). With this new created state a reward is gained, \(R_{t+1}\). This sequence continues until the terminal state is reached, which can be user defined. The agent will set actions such that it maximises its total expected reward, \(\max_{{\pi}} \mathbb{E_{\tau \sim \pi}}\left [ \sum_{t=0}^{\infty} \gamma^{t}R \right ]\).

\newpage
\section{Soft Actor Critic}
\begin{figure}[H]
    \centering
    \includegraphics{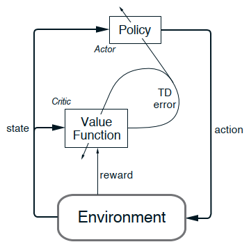}
    \caption{Actor Critic architecture \cite{Sutton2018ReinforcementIntroduction}}
    \label{AC_architecture}
\end{figure}

Actor critic agents have been created to benefit from both policy optimisation and Q-learning methods. The key characteristic of an actor-critic algorithm is that the agent is composed of 2 “sub-agents”, an actor and a critic, figure \ref{AC_architecture}. An actor is learning a policy, \(\pi \), which performs actions maximising the future expected reward.

To maximise exploration by the agent a soft actor critic (SAC) algorithm was implemented. With SAC, the RL objective is modified such that is tries to maximise the expected return and entropy of the agent, \(\max_{{\pi}} \mathbb{E_{\tau \sim \pi}}\left [ \sum_{t=0}^{\infty} \gamma^{t}(R+\alpha H(P)) \right ]\). The temperature parameter \(\alpha\) determines the contribution of H in the agent’s objective where H is the entropy which is a measure of how random the actions of the agent are. Both the actor and critic are represented by a feed-forward neural network and are parameterised by \(\phi\) and \(\theta\) respectively.

The network of the actor takes in the state parameters and outputs a mean, \(\mu_\phi\), and standard deviation, \(\sigma_\phi\), over the action-space of each action. Parameters of the network are updated with the aim to minimise the KL-divergence, equation \ref{kl_div}. \(Z^{\pi_{old}}(s_t)\) is a normalisation factor \cite{Haarnoja2018SoftApplicationsb}.

\begin{equation}
\label{kl_div}
\pi_{new}=\arg\min_{\pi' \in \Pi}D_{KL}\left ( \pi'(\cdot|s_t) || \frac{\exp(\frac{1}{\alpha}Q^{\pi_{old}}(s_t,\cdot))}{Z^{\pi_{old}}(s_t)}\right )
\end{equation}

Minimising the KL-divergence is done by updating the network parameters via stochastic gradient decent, equation \ref{obj_actor} \cite{Sutton2018ReinforcementIntroduction}.

\begin{equation}
    J_{\pi}(\phi)=\mathbb{E}_{s_t \sim D, a_t \sim \pi_{\phi}}\left [ \alpha \log(\pi_{\phi}(a_t|s_t))-Q_{\theta}(s_t,a_t)-Z^{\pi_{old}}(s_t) \right ]
    \label{obj_actor}
\end{equation}

Estimating the gradient with respect to \(\phi\) is not possible, since the expectation depends on \(\phi\). Therefore the reparameterization trick is applied, such that \(a_t=f_{\phi}(s_t,\epsilon_t)=\tanh(\mu_{\phi}+\sigma_{\phi}\odot \epsilon)\), where \(\epsilon\) is a noise vector sampled from a normal distribution, \(\epsilon \sim \mathcal{N}(0,1) \) equation \ref{actor_grad}.  

\begin{equation}
        J_{\pi}(\phi)=\mathbb{E}_{s_t \sim D, \epsilon \sim \mathcal{N}}\left [ \alpha \log(\pi_{\phi}(f_\phi(\epsilon_t,s_t)|s_t))-Q_{\theta}(s_t,f_\phi(\epsilon_t,s_t)) \right ]
        \label{actor_grad}
\end{equation}

\(Z^{\pi_{old}}(s_t)\) is neglected, since it is independent of \(\phi\) and is regarded a constant when estimating the gradient with respect to \(\phi\). The critics’ aim is to learn an optimal soft Q function, equation \ref{q_function}, by minimising the mean squared error, equation \ref{critic_grad}.

\begin{equation}
    Q^{\pi}(s_t,a_t)=\mathbb{E}_{(s_t,a_t)\sim\mathcal{D}}\left [ r+\gamma(Q_{\theta}(s_{t+1},a_{t+1})-\alpha\log\pi(a_{t+1}|s_{t+1})) \right ]
    \label{q_function}
\end{equation}

\begin{equation}
    Q^{\pi}(s_t,a_t)=\mathbb{E}_{(s_t,a_t)\sim\mathcal{D}}\left [ \frac{1}{2}(Q_{\theta}(s_t,a_t)-(r+\gamma\mathbb{E}_{s_{t+1}\sim p}[Q_{\bar{\theta}}(s_{t+1},a_{t+1})-\alpha\log\pi(a_{t+1}|s_{t+1})]))^2 \right ]
    \label{critic_grad}
\end{equation}

Updates of the critic network is performed by applying stochastic gradient descent following equation \ref{Critic_update}.

\begin{equation}
    \widehat{\nabla_{\theta}}J_{Q}(\theta)=\nabla_{\theta}Q_{\theta}\left ( Q_{\theta}(s_t,a_t)-(r+\gamma(Q_{\overline{\theta}}(s_{t+1},a_{t+1})-\alpha \log(\pi_{\phi}(a_{t+1}|s_{t+1})))) \right )
    \label{Critic_update}
\end{equation}

$Q_{\overline{\theta}}$ in equation \ref{Critic_update} is the Q value of a target Q network. This target network is introduced to improve stability of the critic and its parameters are updated by Polyak averaging, ${\overline{\theta}=\tau \theta + (1-\tau) \overline{\theta}}$. $\tau$ is the target smoothing coefficient, which makes for soft target parameter updates.

To determine the relative importance between the reward and entropy the temperature \(\alpha\) is introduced. This parameter determines the stochasticity of the policy and is updated via a multi-constrained objective given in equation \ref{alpha_constraints}, where \(\overline{H}\) is a set target entropy.

\begin{equation}
    \left\{\begin{array}{@{}l@{}}
        \pi^{*}=\arg\max\sum_{t}^{} \mathbb{E}_{(s_t,a_t)\sim\pi}[r(s_t,a_t)]\\
        s.t. \mathbb{E}_{(s_t,a_t)\sim\pi}[-\log\pi(a_t|s_t)]\geq \overline{H}, \forall t
    \end{array}\right.
\label{alpha_constraints}
\end{equation}

The target entropy was fixed, whilst \(\alpha\) was automatically tuned via the following objective function \ref{alpha_objective}.

\begin{equation}
    J(\alpha) = \mathbb{E}_{a_t\sim\pi_{t}}\left [ -\alpha\log\pi_t(a_t|s_t)-\alpha\overline{H} \right ]
    \label{alpha_objective}
\end{equation}

\section{Case description}
For this research, a hydrocarbon mixture was considered for the feed stream with the goal to separate it in individual components with a minimum purity of 95 mol\%. The feed was processed at 12.400 kmol/h with the composition stated in appendix \ref{feed_stream}. The agents' reward was defined as 

\begin{equation}
    r(s_t,a_t)=revenue_{top} + revenue_{bottom} - TAC - penalty
\end{equation}

where revenue was only generated when the purity specification was 95 mol\%. The Total Annualized Cost (TAC) was based on the total capital expendatures (CAPEX) and operational expedatures (OPEX) of the distillation sequence. Appendix \ref{eqn} shows an overview of the design and TAC calculations. A set of penalties were defined to discourage non-sensible column designs, see appendix \ref{agent_penalties}. The state of the environment was defined as the stream conditions, such as temperature, pressure, molar flows, and revenue. The agent's actionspace was limited to setting the number of stages, feed stage, condenser pressure, reflux -and reboil ratio. Boundaries of these parameters were set and are shown in appendix \ref{boundaries}.
To keep track of all streams, a stream table was created which stores all data of a stream if the molar flow is greater or equal to 3.6 kmol/h and the purity specification is not met. Molar flows smaller then 3.6 kmol/h was regarded an outlet and was neither added to the stream table nor was it sold. To decide which stream to separate next was based on the largest molar flow present in the stream table.
In order to compare the design generated by the RL agent, two base cases were created by use of Aspen Plus with optimisation via design specs and a random agent was programmed to take random actions.
The settings used to run the SAC agent are shown in appendix 
\newpage
\section{Case results}
\subsection{Base case}
For the base cases two separation sequences were considered, direct separation or a tree-like sequence. Both these setups were able of generating profit without generating any waste streams. 

For the linear case, four columns were put in series to separate the feed stream in its respective component. Appendix \ref{linear} shows a schematic of the separation train including the column specifications. Each column was able to produce a top and/or bottom stream with an equal or greater molar purity than 95 mol\%. This result was gained with a stages range of 21 to 77 and a diameter ranging from 4.9 to 10.4 meters. Optimised values for the reflux ratio and distillate to feed ratio ranged from 2.72 to 11.28, and 0.35 to 0.67, respectively. The TAC of the linear separation train was 107 M€ and a yearly revenue of 1,552 M€. This results in a yearly profit of 1,445 M€.

In the case of a tree-like separation sequence two columns were put in series and the last two columns were put in parallel. This configuration allows for parallel processing of the top and bottom stream of the second column. \\ Appendix \ref{tree} shows the final column configuration with the columns respective dimensions, TAC and revenue. The range of diameter and number of stages in this configuration is from 6.5 to 9.5 meter and 21 to 85, respectively. Optimisation of the columns was done with the build-in design spec option of Aspen. For optimisation the distillate to feed ratio and reflux ratio were chosen, since this directly impacts the desired product, the distillate. The optimised values reflux ratio and distillate to feed ratio ranged from 1.09 to 8.56 and 0.34 to 0.77, respectively. A tree-like separation setup had a TAC of 107 M€, with a yearly revenue of 1,505 M€. This yields a profit of 1,398 M€.


\subsection{SAC agent}
For the SAC agent 2 cases will be evaluated, a maximum profit and a best average case.
\subsubsection{Max profit}
\begin{figure}[!ht]
    \centering
    \includegraphics[scale=0.6]{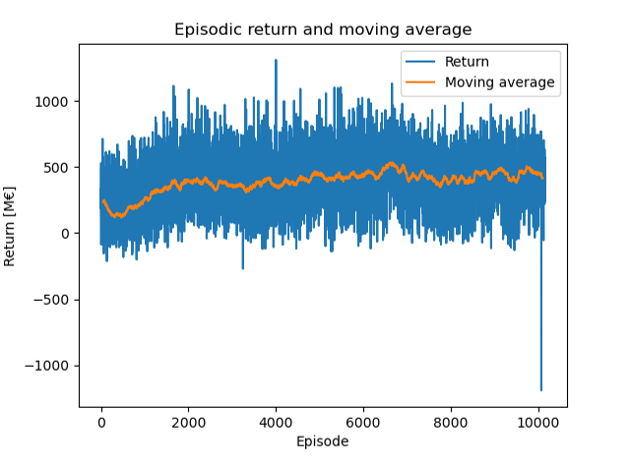}
    \caption{Return per episode and moving average of RL agent yielding the highest profit}
    \label{fig:return_max_profit}
\end{figure}

Utilising the SAC agent with the settings provided in appendix \ref{agent_params} generated a maximum profit of 1,311 M€ and an average profit of 384 M€. The maximum profit was generated by utilising 8 columns, which are shown in appendix \ref{app:max_profit}. Dimensions of the columns vary between 21 and 92 number of stages and a diameter ranging from 5 to 9.9 meters. All open streams in this configuration are at spec besides the bottom stream of column 4. This stream is not considered an outlet, since the molar flow is greater then 3.6 kmol/h. Hence the agent would continue separating this stream if it was allowed to. Figure \ref{fig:return_max_profit} shows the trend in return of the RL agent. Here can be seen that there is almost no increase in return after 3000 episodes.

\newpage

\begin{figure}[!ht]
    \centering
    \includegraphics[scale=0.6]{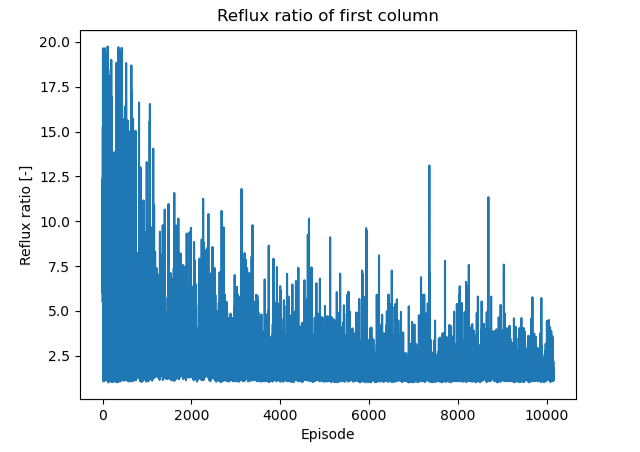}
    \caption{Reflux ratio of first column of each episode}
    \label{fig:rr_max_profit}
\end{figure}

Observing the agents' actions for the first column provides a sense of how the agent develops over time, because for the first column the input stream is always the same. Figure \ref{fig:rr_max_profit} shows a converging trend for the reflux ratio over episodes. This is indicative for the agent adjusting the parameters to converge to an optimum. Appendix \ref{app:max_profit} shows the critic loss and entropy of the agent. These plots together with the figure \ref{fig:rr_max_profit} indicate a fast learning agent which takes little time to explore and start exploiting actions early on in the process. Hence it would be desired to decrease the rate at which the agent learns in order to provide sufficient time to explore new configurations.

\subsubsection{Best average}
\begin{figure}[H]
    \centering
    \includegraphics[scale=0.6]{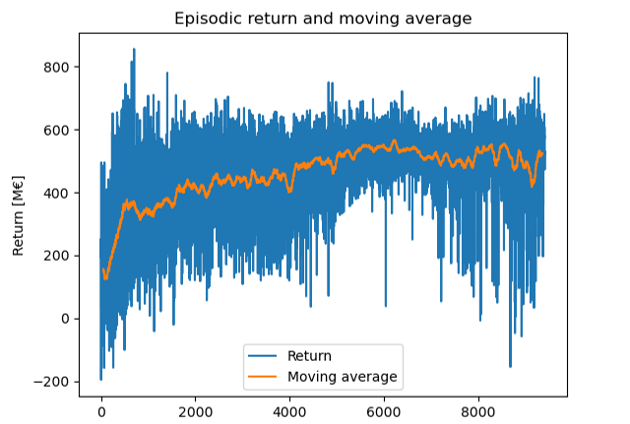}
    \caption{Return per episode and a moving average of the RL agent yielding the best average}
    \label{fig:my_label}
\end{figure}

Increasing performance based on the average profit was found with increasing the number of agent updates per episode to 4 and keeping the other parameters similar to appendix \ref{agent_params}. This adjustment yielded an average profit of 457 M€ and a maximum profit of 856 M€ within 9405 episodes. Figure \ref{fig:my_label} shows the trend in return and moving average. At first the profit increases sharp, but decreases as it continues. In comparison to the max profit case shows the adjustment a steeper increase overall. This increase is observed, since the agent observes the replay buffer more often. From figure \ref{fig:my_label} can be seen that the agent converges towards until the 6500th episode. After this episode it starts exploring again, hence the increased variance over returns. This trend was also observed in the set column parameters. 

\begin{figure}[!ht]
    \centering
    \includegraphics[scale=0.6]{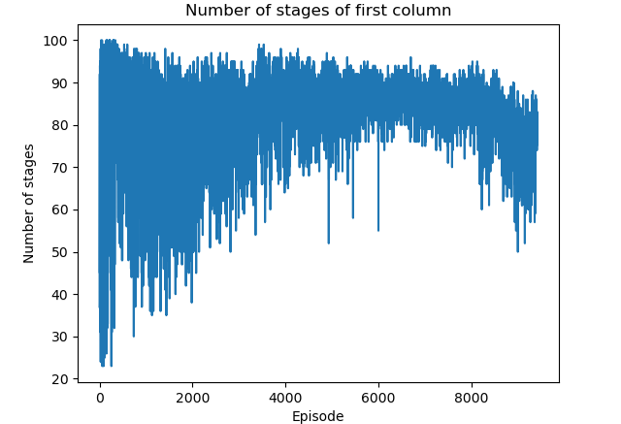}
    \caption{Number of stage of the first column of each episode}
    \label{fig:number of stages best average}
\end{figure}

Figure \ref{fig:number of stages best average} shows the number of stages in the first column of each episode. It shows the same converging trend until the 6500th episode after which it start setting new values and gaining new experiences.

\subsection{Overall}
The stability and speed were tracked and are presented in table \ref{table:conv_stats}. From this table can be seen that the RL agent almost always chooses to separate the stream, which means that it takes every opportunity to gain experience. Another good result is the low percentage of convergence error of the flowsheet in Aspen. This indicates that the agent almost always takes actions which are solvable by the flowsheet. However, the implementation of RL in commercial simulation software showed some pitfalls with regard to speed and stability. It took the agent 193 hours to complete 10,151 episodes. This extensive time in combination with the unstable connection which rises RPC errors makes for a tedious way of performing a sensitivity analysis of the agents' hyperparameters.

\begin{table}[!ht]

\centering
\caption{Convergence statistics }
\label{table:conv_stats}
\begin{tabular}{lcccc} 
\toprule
\multicolumn{1}{c}{\multirow{2}{*}{\textbf{Agent}}} & \multirow{2}{*}{\textbf{Episodes}}              & \multirow{2}{*}{\textbf{Duration}}               & \multicolumn{2}{c}{\textbf{Separate}}                                                                \\
\multicolumn{1}{c}{}                                &                                                 &                                                  & \textbf{Yes}                                     & \textbf{No}                                       \\ 
\hline
\textbf{RL agent, standard}                         & 10,151                                          & 193:43:13                                        & 99.8\%                                           & 0.2\%                                             \\
\textbf{RL agent, 4 agent updates}                  & 9405                                            & 160:58:43                                        & 98.9\%                                           & 1.1\%                                             \\ 
\hline
                                                    & \multicolumn{1}{l}{}                            & \multicolumn{1}{l}{}                             & \multicolumn{1}{l}{}                             & \multicolumn{1}{l}{}                              \\ 
\hline
                                                    & \multicolumn{4}{c}{\textbf{Aspen runs}}                                                                                                                                                                   \\
\multicolumn{1}{c}{\multirow{2}{*}{\textbf{Agent}}} & \multirow{2}{*}{\textbf{Lost contact}}          & \multirow{2}{*}{\textbf{Convergence error}}      & \multicolumn{2}{c}{\textbf{Converged }}                                                              \\
\multicolumn{1}{c}{}                                &                                                 &                                                  & \textbf{With warnings}                           & \textbf{Without warnings}                         \\ 
\hline
\textbf{RL agent, standard}                         & 0.2\%                                           & 0.01\%                                           & 23.8\%                                           & 76.0\%                                            \\
\textbf{RL agent, 4 agent updates}                  & \begin{tabular}[c]{@{}c@{}}0.2\%\\\end{tabular} & \begin{tabular}[c]{@{}c@{}}0.03\%\\\end{tabular} & \begin{tabular}[c]{@{}c@{}}35.4\%\\\end{tabular} & \begin{tabular}[c]{@{}c@{}}64.4\%\\\end{tabular}  \\
\bottomrule
\end{tabular}
\end{table}

\newpage
\section{Conclusion and future}
This paper showed the applicability of RL in a chemical engineering context. However, the RL agent was not able to outperform the created base case. To potentially outperform the base case it is necessary to increase insight on the effect of the hyperparameters. Based on the result in this study, the agent stops exploring too early, therefore the learning of the agent should be slowed down. Parameters which influence this learning are the learning rate of $\alpha$ and the target smoothing parameter of the Q-target network. Lowering these will lead to an increased learning time, but will allow for more exploration by the agent.

To speed up the convergence to optimal solutions, good estimates can be passed on to the agent by first simulating the column with a DSTWU in Aspen Plus and feed the DSTWU column design to the SAC agent as initialisation.

The connection between Python and Aspen Plus was experienced as unstable and Aspen Plus acts as a rather slow RL environment. Therefore it would be advised to create a distillation model in Python, such that no external communication is required.

\bibliographystyle{unsrt}  
\bibliography{references}  

\begin{thebibliography}{10}

\bibitem{SilverASelf-playb}
David Silver, Thomas Hubert, Julian Schrittwieser, Ioannis Antonoglou, Matthew
  Lai, Arthur Guez, Marc Lanctot, Laurent Sifre, Dharshan Kumaran, Thore
  Graepel, Timothy Lillicrap, Karen Simonyan, and Demis Hassabis.
\newblock {A general reinforcement learning algorithm that masters chess, shogi
  and Go through self-play}.

\bibitem{Hoskins1992ProcessLearning}
J.~C. Hoskins and D.~M. Himmelblau.
\newblock {Process control via artificial neural networks and reinforcement
  learning}.
\newblock {\em Computers {\&} Chemical Engineering}, 16(4):241--251, 4 1992.

\bibitem{Shin2019ReinforcementControl}
Joohyun Shin, Thomas~A. Badgwell, Kuang~Hung Liu, and Jay~H. Lee.
\newblock {Reinforcement Learning – Overview of recent progress and
  implications for process control}.
\newblock {\em Computers {\&} Chemical Engineering}, 127:282--294, 8 2019.

\bibitem{Midgley2019ReinforcementSynthesis}
Laurence Midgley and Michael Thomson.
\newblock {Reinforcement learning for chemical engineering process synthesis}.
\newblock {\em Bachelor Thesis, University of Cape Town}, 11 2019.

\bibitem{Midgley2020DeepSynthesisb}
Laurence~Illing Midgley.
\newblock {Deep Reinforcement Learning for Process Synthesis}.
\newblock {\em https://arxiv.org/abs/2009.13265v1}, 9 2020.

\bibitem{Stops2022FlowsheetNetworksb}
Laura Stops, Roel Leenhouts, Qinghe Gao, and Artur~M. Schweidtmann.
\newblock {Flowsheet synthesis through hierarchical reinforcement learning and
  graph neural networks}.
\newblock {\em https://arxiv.org/abs/2207.12051}, 7 2022.

\bibitem{Gottl2021AutomatedConcept}
Quirin G{\"{o}}ttl, Yannic T{\"{o}}nges, Dominik~G. Grimm, and Jakob Burger.
\newblock {Automated Flowsheet Synthesis Using Hierarchical Reinforcement
  Learning: Proof of Concept}.
\newblock {\em Chemie Ingenieur Technik}, 93(12):2010--2018, 12 2021.

\bibitem{Gottl2022AutomatedLearning}
Quirin G{\"{o}}ttl, Dominik~G. Grimm, and Jakob Burger.
\newblock {Automated synthesis of steady-state continuous processes using
  reinforcement learning}.
\newblock {\em Frontiers of Chemical Science and Engineering}, 16(2):288--302,
  2 2022.

\bibitem{Sutton2018ReinforcementIntroduction}
Richard~S. Sutton and Andrew~G. Barto.
\newblock {\em {Reinforcement Learning: An Introduction}}.
\newblock MIT Press, Cambridge, second edition, 2018.

\bibitem{Haarnoja2018SoftApplicationsb}
Tuomas Haarnoja, Aurick Zhou, Kristian Hartikainen, George Tucker, Sehoon Ha,
  Jie Tan, Vikash Kumar, Henry Zhu, Abhishek Gupta, Pieter Abbeel, and Sergey
  Levine.
\newblock {Soft Actor-Critic Algorithms and Applications}.
\newblock {\em https://arxiv.org/abs/1812.05905}, 12 2018.

\end{thebibliography}
\newpage
\section*{Appendix}

\begin{appendices}

\section{Feed stream}
\begin{table}[ht]
\renewcommand{\arraystretch}{1.5}
\centering
\caption{Feed stream specifications}
\label{feed_stream}
\begin{tabular}{lcl}
\hline
Component   & \multicolumn{2}{c}{Mol\%}               \\ \hline
$C_{2}H_{6}$ & \multicolumn{2}{c}{0.06}                \\
$C_{3}H_{8}$  &\multicolumn{2}{c}{33.69}               \\
i$C_{4}H_{10}$ & \multicolumn{2}{c}{35.65} \\
n$C_{4}H_{10}$     & \multicolumn{2}{c}{15.32}               \\
i$C_{5}H_{12}$      & \multicolumn{2}{c}{10.21}               \\
n$C_{5}H_{12}$     & \multicolumn{2}{c}{5.13}                \\ \hline
Molar flow  & [kmol/h]                   & 12,400 \\
Temperature & [$^{\circ}C$] & 105    \\
Pressure    & [bar] & 17     \\ \hline
\end{tabular}
\end{table}

\section{Agent penalties}
\label{agent_penalties}
\begin{table}[ht]
\renewcommand{\arraystretch}{1.5}
\centering
\caption{Agent penalties}
\label{penalties}
\begin{tabular}{cccllll}
\hline
Parameter & Non-sensible condition & Penalty & & & & \\ \hline
Q\(_{condenser}\) & \(\geq\) 0 & 5 M€ & & & & \\
Q\(_{reboiler}\) & \(\leq\) 0 & 5 M€  & & & & \\
T\(_{reboiler}\) & \(\geq\) T\(_{steam}\) & 5 M€ & & & & \\
$\Delta T_{lm_{condenser}}$ & $\leq T_{cool}$  & 5 M€    & \multicolumn{1}{c}{} & \multicolumn{1}{c}{} & \multicolumn{1}{c}{} & \multicolumn{1}{c}{} \\
Flowsheet error & - & 10 M€   & & & & \\ \hline       
\end{tabular}
\end{table}

\section{Action boundaries}
\label{boundaries}
\begin{table}[!ht]
\renewcommand{\arraystretch}{1.4}
\centering
\caption{Parameter boundaries}
\label{bounds}
\begin{tabular}{lccc}
\hline
\multicolumn{1}{c}{\multirow{2}{*}{Parameter}} & \multirow{2}{*}{Type} & \multicolumn{2}{c}{Bound}                \\
\multicolumn{1}{c}{}      &  & Min                & Max                 \\ \hline
Number of stages          & Discrete              & 20                 & 100                 \\
Feed stage                & Discrete              & 0.2*n\(_{stages}\) & 0.8*n\(_{stages}\)  \\
Condenser pressure        & Real                  & 0.5                & 30                  \\
Reflux ratio              & Real                  & \multirow{2}{*}{1} & \multirow{2}{*}{20} \\
Boilup ratio              & Real                  &                    &                     \\ \hline
\end{tabular}
\end{table}
\newpage

\section{Equations}
\label{eqn}

\begin{table}[!ht]
\renewcommand{\arraystretch}{1.4}
\centering
\caption{Design and economical equations}
\label{eqns}
\begin{tabular}{lclc}
Design equations& &                     &                        \\
\multicolumn{4}{c}{\textbf{Column diameter}}  \\
\multicolumn{4}{c}{$D=1.1*\max(D_{eff})$}     \\
Where           & &                     &                        \\
\multicolumn{1}{l}{$D$}            & =                         & \multicolumn{1}{l}{diameter}           & {[}$m${]}              \\
\multicolumn{1}{l}{$D_{eff}$}      & =                         & \multicolumn{1}{l}{effective diameter} & {[}$m${]}              \\
\multicolumn{1}{l}{$n_s$}          & =                         & \multicolumn{1}{c}{0.2 \textless{} $n_s$ \textless 0.8}          & {[}$-${]}                \\
                & &                     &                        \\
\multicolumn{4}{c}{$D_{eff}=\sqrt{\frac{4*\Phi_{mol,n_s}^{vapor}}{3.1416*f}*\sqrt{\frac{R*T_s*Mw_{n_s}^{vapor}*1000}{P}}}$} \\
$\Phi_{mol,n_s}^{vapor}$                         & =                         & molar vapor flow rate                  & {[}$\frac{mol}{s}${]}  \\
$f$             & =                         & gas load factor (fixed, 1.6)           & {[}$Pa^{0.5}${]}       \\
$R$             & =                         & molar gas factor (fixed, 1.6)          & {[}$\frac{J}{mol*K}${]}\\
$T_{n_s}$       & =                         & temperature         & {[}$K${]}              \\
$Mw_{n_s}^{vapor}$                    & =                         & vapor molar weight  & {[}$\frac{g}{mol}${]}  \\
P               & =                         & column pressure     & {[}$Pa${]}             \\& &    &\\
\multicolumn{4}{c}{\textbf{Column height}}    \\
\multicolumn{4}{c}{$L=n_s*HETP+H_0$}\\
Where & & & \\
L & = & column height       & {[}$m${]} \\
HETP & = & height equivalent to a theoretical plate & {[}$m${]} \\
$H_0$ & = & clearance (fixed, 4)& {[}$m${]} \\
& & & \\
\multicolumn{4}{c}{\textbf{Condenser area}}   \\
\multicolumn{4}{c}{$A_{cnd}=-\frac{Q_{cnd}}{K_{cnd}*\Delta T_{lm,cnd}}$}\\
Where & & & \\
$A_{cnd}$ & = & condenser area      & {[}$m^2${]} \\
$Q_{cnd}$ & = & condenser duty, negative value & {[}$W${]} \\
$K_{cnd}$ & = & heat transfer coefficient (fixed, 500) & {[}$\frac{W}{m^2 * ^{\circ}C}${]} \\
$\Delta T_{lm,cnd}$ & = & logarithmic mean temperature difference & {[}\textdegree$C${]} \\
& & & \\
\multicolumn{4}{c}{$\Delta T_{lm,cnd}=\sqrt[3]{\frac{(T_{cnd}-T_{in}^{cool})*(T_{cnd}-T_{out}^{cool})*((T_{cnd}-T_{in}^{cool})+(T_{cnd}-T_{out}^{cool}))}{2}}$}                             \\
Where           &                           &                     &                                                  \\
$T_{cnd}$       & =                         & condenser temperature                                                                      & [\textdegree $C$]                                \\
$T_{in}^{cool}$ & =                         & supply temperature cooling water (fixed, 30)                                               & {[}\textdegree$ C${]}                                \\
$T_{out}^{cool}$& =                         & return temperature cooling water (fixed, 40)                                               & {[}\textdegree$C${]}                                \\
                &                           &                     &                                                  \\
\end{tabular}
\end{table}

\begin{table}[H]
\renewcommand{\arraystretch}{1.4}
\centering
\caption*{}
\label{eqns_economical}
\begin{tabular}{lclc}
\multicolumn{4}{c}{\textbf{Reboiler area}}    \\
\multicolumn{4}{c}{$A_{rbl}=\frac{Q_{rbl}}{K_{rbl}*\Delta T_{lm,rbl}}$}                                              \\
Where           &                           &                     &                                                  \\
$A_{rbl}$       & =                         & reboiler area       & {[}$m^2${]}                                      \\
$Q_{rbl}$       & =                         & reboiler duty, positive value                                                              & {[}$W${]}                                        \\
$K_{rbl}$       & =                         & heat transfer coefficient (fixed, 800)                                                     & {[}$\frac{W}{m^2 * ^{\circ}C}${]}                         \\
$\Delta T_{lm,rbl}$                                                                    & =                         & temperature difference                                                                     & {[}\textdegree$C${]}                                        \\
                &                           &                     &                                                  \\
\multicolumn{4}{c}{$\Delta T_{lm,rbl} = T_{steam}-T_{rbl}$}                                                          \\
Where           &                           &                     &                                                  \\
$T_{steam}$    & =                         & steam temperature (fixed, 201)                                                             & {[}\textdegree$ C${]}                                \\
$T_{rbl}$      & =                         & reboiler temperature& {[}\textdegree$ C${]}                                \\
                &                           &                     &     \\
                                               &                           &                     &                                                  \\
\multicolumn{4}{c}{\textbf{Investment cost}}  \\
\multicolumn{4}{c}{$I=F_{cap}*C_{inv}$}       \\
Where           &                           &                     &                                                  \\
$I$             & =                         & investment          & {[}$M$\texteuro{]}                                   \\
$F_{cap}$       & =                         & capital charge factor                                                                      & {[}$-${]}                                        \\
$C_{inv}$       & =                         & investment cost     & {[}$M$\texteuro{]}                                   \\
                &                           &                     &                                                  \\
\multicolumn{4}{c}{$C_{inv}=F_L*C_{eqp}$} \\
$F_L$           & =                         & Lang factor (fixed, 5)                                                                     & {[}$-${]}                                        \\
                &                           &                     &                                                  \\
\multicolumn{4}{c}{\textbf{Equipment cost}}   \\
\multicolumn{4}{c}{$C_{eqp}=\frac{C_{col}+C_{int}+C_{cnd}+C_{rbl}}{1000000}$}                                        \\
$C_{eqp}$       & =                         & equipment cost      & {[}$M$\texteuro{]}                                   \\
$C_{col}$       & =                         & column cost         & {[}\texteuro{]}                                    \\
$C_{int}$       & =                         & internals cost      & {[}\texteuro{]}                                    \\
$C_{cnd}$       & =                         & condenser cost      & {[}\texteuro{]}                                    \\
$C_{rbl}$       & =                         & reboiler cost       & {[}\texteuro{]}                                    \\
                &                           &                     &                                                  \\
\multicolumn{4}{c}{\textbf{Column cost}}      \\
\multicolumn{4}{c}{$C_{col}=\frac{0.9*Ms_{2018}*937.64*D^{1.066}*L^{0.802}*F_{c,col}}{280}$}                         \\
$C_{col}$       & =                         & column cost         & {[}\texteuro{]}                                    \\
$Ms_{2018}$     & =                         & Marshall \& Swift equipment index 2018 (fixed, 1638.2)                                     & {[}$-${]}                                          \\
$F_{c,col}$     & =                         & correction factor for column                                                               & {[}$-${]}                                          \\
                &                           &                     &                                                  \\

\end{tabular}
\end{table}

\newpage

\begin{table}[H]
\renewcommand{\arraystretch}{1.5}
\centering
\caption*{}
\label{eqns3_economical}
\begin{tabular}{lclc}
\multicolumn{4}{c}{$F_{c,col}=F_m+F_p$}       \\
$F_{m}$        & =                         & correction factor for column shell material (fixed, 1)                                     & {[}$-${]}                                          \\
$F_p$           & =                         & correction factor for column pressure (fixed, 1)                                           & {[}$-${]}                                          \\
                &                           &                     &                                                  \\
\multicolumn{4}{c}{\textbf{Internal costs}}   \\
\multicolumn{4}{c}{$C_{int}=\frac{0.9*Ms_{2018}*97.24*D^{1.55}*L*F_{c,int}}{280}$}                                   \\
$C_{int}$       & =                         & internal cost       & [\texteuro]                                    \\
$F_{c,int}$     & =                         & correction factor for internals                                                            & {[}$-${]}                                        \\
                &                           &                     &                                                  \\
\multicolumn{4}{c}{$F_{int,c}=F_{int,s}+F_{int,t}+F_{int,m}$}                                                        \\
$F_{int,s}$     & =                         & correction factor for tray spacing (fixed, 1.4)                                            & {[}$-${]}                                        \\
$F_{int,t}$     & =                         & correction factor for tray type (fixed, 0)                                                 & {[}$-${]}                                        \\
$F_{int,m}$     & =                         & correction factor for internals material (fixed, 0)                                        & {[}$-${]}                                        \\

\multicolumn{4}{c}{\textbf{Condenser cost}}   \\
\multicolumn{4}{c}{$C_{cnd}=\frac{0.9*Ms_{2018}*474.67*A_{cnd}^{0.65}*F_{c,cnd}}{280}$}                              \\
$C_{cnd}$       & =                         & condenser cost      & [\texteuro]                                    \\
$F_{cnd,c}$     & =                         & correction factor for condenser                                                            & {[}$-${]}                                        \\
                &                           &                     &                                                  \\
\multicolumn{4}{c}{$F_{cnd,c}=(F_{htx,P}+F_{htx.d})*F_{htx,m}$}                                                      \\
$F_{htx,P}$     & =                         & correction factor for pressure (fixed, 0)                                                  & {[}$-${]}                                        \\
$F_{htx.d}$     & =                         & correction factor for design type: fixed - tube sheet (fixed, 0.8)                         & {[}$-${]}                                        \\
$F_{htx,m}$     & =                         & correction factor for material (fixed, 1)                                                  & {[}$-${]}                                        \\
                &                           &                     &                                                  \\
\multicolumn{4}{c}{\textbf{Reboiler cost}}    \\
\multicolumn{4}{c}{$C_{rbl}=\frac{0.9*Ms_{2018}*474.67*A_{rbl}^{0.65}*F_{c,rbl}}{280}$}                              \\
$F_{rbl,c}$     & =                         & condenser duty, negative value                                                             & {[}$W${]}                                        \\
                &                           &                     &                                                  \\
\multicolumn{4}{c}{$F_{rbl,c}=(F_{htx,P}+F_{htx,d})*F_{htx,m}$}                                                      \\
                &                           &                     &                                                  \\
\multicolumn{4}{c}{\textbf{Condenser operation cost}}                                                                \\
\multicolumn{4}{c}{$C_{cnd,operation}=\frac{E_{cnd}*E_{cost,cnd}}{1,000,000}$}                                       \\
$C_{cnd,operation}$                                                                    & =                         & operational cost condenser                                                                 & {[}$M$\texteuro{]}                                   \\
$E_{cnd}$       & =                         & condenser energy    & {[}$\frac{kWh}{year}${]}                         \\
$E_{cost,cnd}$  & =                         & energy price        & {[}$\frac{\euro}{kWh}${]}                        \\
                &                           &                     &                                                  \\
\multicolumn{4}{c}{$E_{cnd}=\frac{uptime*Q_{cnd}}{1000}$}                                                            \\
\multicolumn{4}{c}{$E_{cost,cnd}=6*10^{-6}*T_{cool}^{2}-0.0006*T_{cool}+0.0163$}                                     \\
                &                           &                     &                                                  \\
\end{tabular}
\end{table}
\newpage
\begin{table}[!ht]
\renewcommand{\arraystretch}{1.5}
\centering
\caption*{}
\label{eqns2_economical}
\begin{tabular}{lclc}
\multicolumn{4}{c}{\textbf{Reboiler operation cost}}                                                                 \\
\multicolumn{4}{c}{$C_{rbl,operation}=\frac{Q_{rbl}*M*c_{steam}*3600*uptime}{\Delta H_{vap}*1,000,000}$}             \\
$C_{rbl,operation}$                                                                    & =                         & operational costs reboiler                                                                 & {[}$M$\texteuro{]}                                   \\
$M$             & =                         & molar weight of water (fixed, 18)                                                          & {[}$\frac{g}{mol}${]}                            \\
$c_{steam}$     & =                         & steam price at 16 bar (fixed, 18)                                                          & [$\frac{\text{\euro}}{ton}$]                        \\
$\Delta H_{vap}$ & =                         & molar heat of vaporization of 16 bar steam (fixed, 34794)                                  & {[}$\frac{J}{mol}${]}

\end{tabular}
\end{table}

\section{Reinforcement learning agent parameters}
\label{agent_params}
\begin{table}[!ht]
\centering
\caption{Reinforcement learning agent parameters}
\begin{tabular}{lcc} 
\toprule
\multicolumn{1}{c}{\textbf{Agent}}   & Max profit & Best average \\ 
\hline
Optimiser & \multicolumn{2}{c}{Adam}           \\
Learning rate & \multicolumn{2}{c}{3e-4}   \\
Discount factor & \multicolumn{2}{c}{0.99}           \\
Temperature & \multicolumn{2}{c}{Adaptive}       \\
Target smoothing coefficient       & \multicolumn{2}{c}{0.005}          \\
Target entropy & \multicolumn{2}{c}{-5 (=-dim[A])}  \\
Replay buffer size & \multicolumn{2}{c}{\begin{tabular}[c]{@{}c@{}}Min. 320,  \\Max. 320.000\end{tabular}}  \\
Reward scale & \multicolumn{2}{c}{10}             \\
Network updates per episode & 1 & 4 \\ 
\hline
\multicolumn{1}{c}{\textbf{Network}} & \multicolumn{1}{l}{~} & \multicolumn{1}{l}{}                 \\ 
\hline
Hidden layers & \multicolumn{2}{c}{2}              \\
Hidden layer size                  & \multicolumn{2}{c}{128}            \\
Type of network                    & \multicolumn{2}{c}{FeedForward}    \\
\bottomrule
\end{tabular}
\end{table}

\newpage

\section{Base case results - Linear}
\label{linear}

\begin{figure}[H]
    \centering
    \includegraphics[angle=90,origin=c]{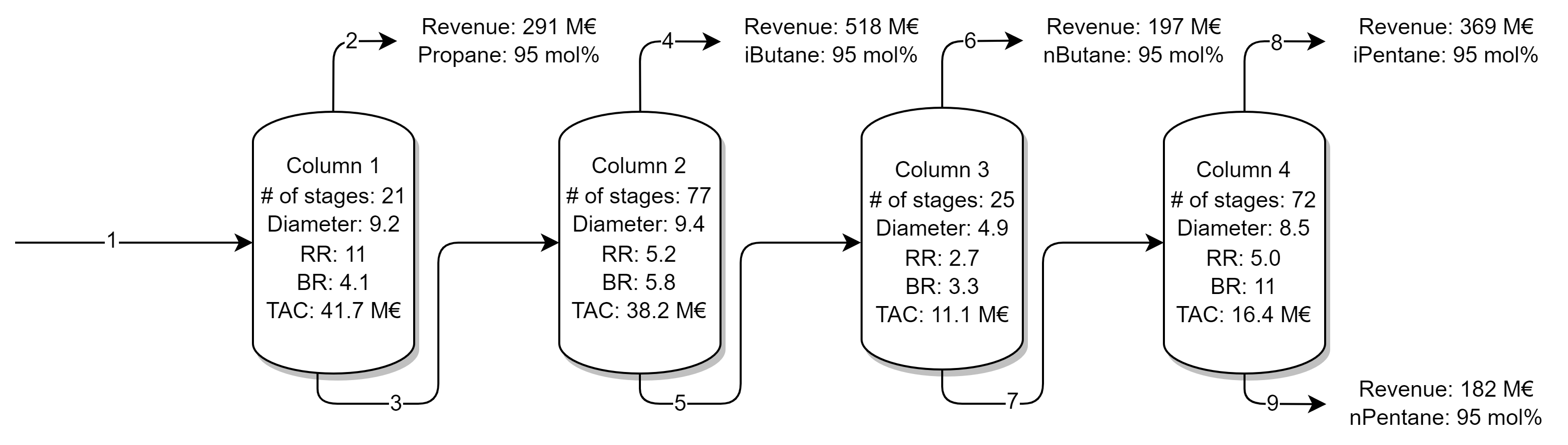}
    \caption{Base case linear separation sequence}
    \label{linear_sequence}
\end{figure}

\begin{table}[H]
\centering
\caption{Linear column specifications with optimised reflux ratio and distillate/feed ratio}
\label{linear_column_spec}
\begin{tabular}{lccccc}
Column             & Unit      & 1     & 2    & 3    & 4     \\
Condenser pressure & {[}bar{]} & 17.2  & 7.2  & 7.2  & 2.0   \\
Reflux ratio       & {[}-{]}   & 11.28 & 5.24 & 2.72 & 5.04  \\
Boilup ratio       & {[}-{]}   & 4.07  & 5.76 & 3.31 & 10.64 \\
Distillate/Feed    & {[}-{]}   & 0.35  & 0.55 & 0.49 & 0.67  \\
Condenser duty     & {[}MW{]}  & -181  & -130 & -34  & -49   \\
Reboiler duty      & {[}MW{]}  & 130   & 114  & 34   & 44   
\end{tabular}
\end{table}

\begin{table}[H]
\centering
\caption{Stream table linear base case}
\begin{tabular}{lccllcccllll} 
\toprule
\multicolumn{12}{c}{\begin{tabular}[c]{@{}c@{}}\textbf{Molar purity}\\\textbf{mol\%}\end{tabular}}          \\ 
\hline
\multicolumn{1}{c}{Stream
  number} & 1 & 2 & \multicolumn{2}{c}{3} & 4 & 5 & 6 & \multicolumn{2}{c}{7} & 8 & 9 \\
C$_2$H$_6$ & 0.1 & 0.1 & \multicolumn{2}{c}{0.0} & 0.0 & 0.0 & 0.0 & \multicolumn{2}{c}{0.0}  & 0.0 & 0.0  \\
C$_3$H$_8$ & 33.2 & 95.0 & \multicolumn{2}{c}{0.5}  & 0.9 & 0.0 & 0.0 & \multicolumn{2}{c}{0.0}  & 0.0 & 0.0 \\
iC$_4$H$_{10}$ & 35.8 & 4.6 & \multicolumn{2}{c}{52.4} & 95.0 & 1.1 & 2.3 & \multicolumn{2}{c}{0.0}  & 0.0 & 0.0 \\
nC$_4$H$_{10}$ & 15.4  & 0.2 & \multicolumn{2}{c}{23.5} & 4.1 & 46.8  & 95.0  & \multicolumn{2}{c}{0.9}  & 1.4 & 0.0 \\
iC$_5$H$_{12}$ & 10.3  & 0.0 & \multicolumn{2}{c}{15.8} & 0.0 & 34.7  & 2.4 & \multicolumn{2}{c}{65.4} & 95.0  & 5.0 \\
nC$_5$H$_{12}$ & 5.2 & 0.0 & \multicolumn{2}{c}{7.9}  & 0.0 & 17.4  & 0.3 & \multicolumn{2}{c}{33.7} & 3.6 & 95.0  \\ 
\hline
\multicolumn{12}{c}{\begin{tabular}[c]{@{}c@{}}\textbf{Molar flow}\\\textbf{kmol/h}\end{tabular}}           \\ 
\hline
Stream number  & 1   & 2   & \multicolumn{2}{c}{3}    & 4   & 5   & 6   & \multicolumn{2}{c}{7}    & \multicolumn{1}{c}{8}    & \multicolumn{1}{c}{9}    \\
C$_2$H$_6$           & 6   & 6   & \multicolumn{2}{c}{0}    & 0   & 0   & 0   & \multicolumn{2}{c}{0}    & \multicolumn{1}{c}{0}    & \multicolumn{1}{c}{0}    \\
C$_3$H$_8$           & 3995  & 3955  & \multicolumn{2}{c}{40}   & 40  & 0   & 0   & \multicolumn{2}{c}{0}    & \multicolumn{1}{c}{0}    & \multicolumn{1}{c}{0}    \\
iC$_4$H$_{10}$         & 4312  & 192 & \multicolumn{2}{c}{4120} & 4079  & 40  & 40  & \multicolumn{2}{c}{0}    & \multicolumn{1}{c}{0}    & \multicolumn{1}{c}{0}    \\
nC$_4$H$_{10}$        & 1857  & 10  & \multicolumn{2}{c}{1847} & 174 & 1673  & 1656  & \multicolumn{2}{c}{17}   & \multicolumn{1}{c}{17}   & \multicolumn{1}{c}{0}    \\
iC$_5$H$_{12}$          & 1240  & 0   & \multicolumn{2}{c}{1240} & 0   & 1240  & 42  & \multicolumn{2}{c}{1198} & \multicolumn{1}{c}{1168} & \multicolumn{1}{c}{30}   \\
nC$_5$H$_{12}$         & 623 & 0   & \multicolumn{2}{c}{623}  & 0   & 623 & 5   & \multicolumn{2}{c}{618}  & \multicolumn{1}{c}{45}   & \multicolumn{1}{c}{574}  \\ 
Total          & 12033 & 4164  & \multicolumn{2}{c}{7869} & 4293  & 3576  & 1743  & \multicolumn{2}{c}{1833} & \multicolumn{1}{c}{1229} & \multicolumn{1}{c}{604}  \\ 
\hline
Temperature [°C] & \multicolumn{1}{l}{105}  & \multicolumn{1}{l}{51.6} & 106  & 66.1              & \multicolumn{1}{l}{52.2} & \multicolumn{1}{l}{80.5} & \multicolumn{1}{l}{65.3} & 102 & 51.8               & 48.6  & 57.8  \\
Pressure [bar] & \multicolumn{1}{l}{17.2} & \multicolumn{1}{l}{17.2} & 17.2 & 7.2               & \multicolumn{1}{l}{7.2}  & \multicolumn{1}{l}{7.2}  & \multicolumn{1}{l}{7.2}  & 7.2 & 2.0                & 2.0 & 2.0 \\
\bottomrule
\end{tabular}
\end{table}

\section{Base case results - Tree}
\label{tree}
\begin{figure}[!htbp]
    \centering
    \includegraphics{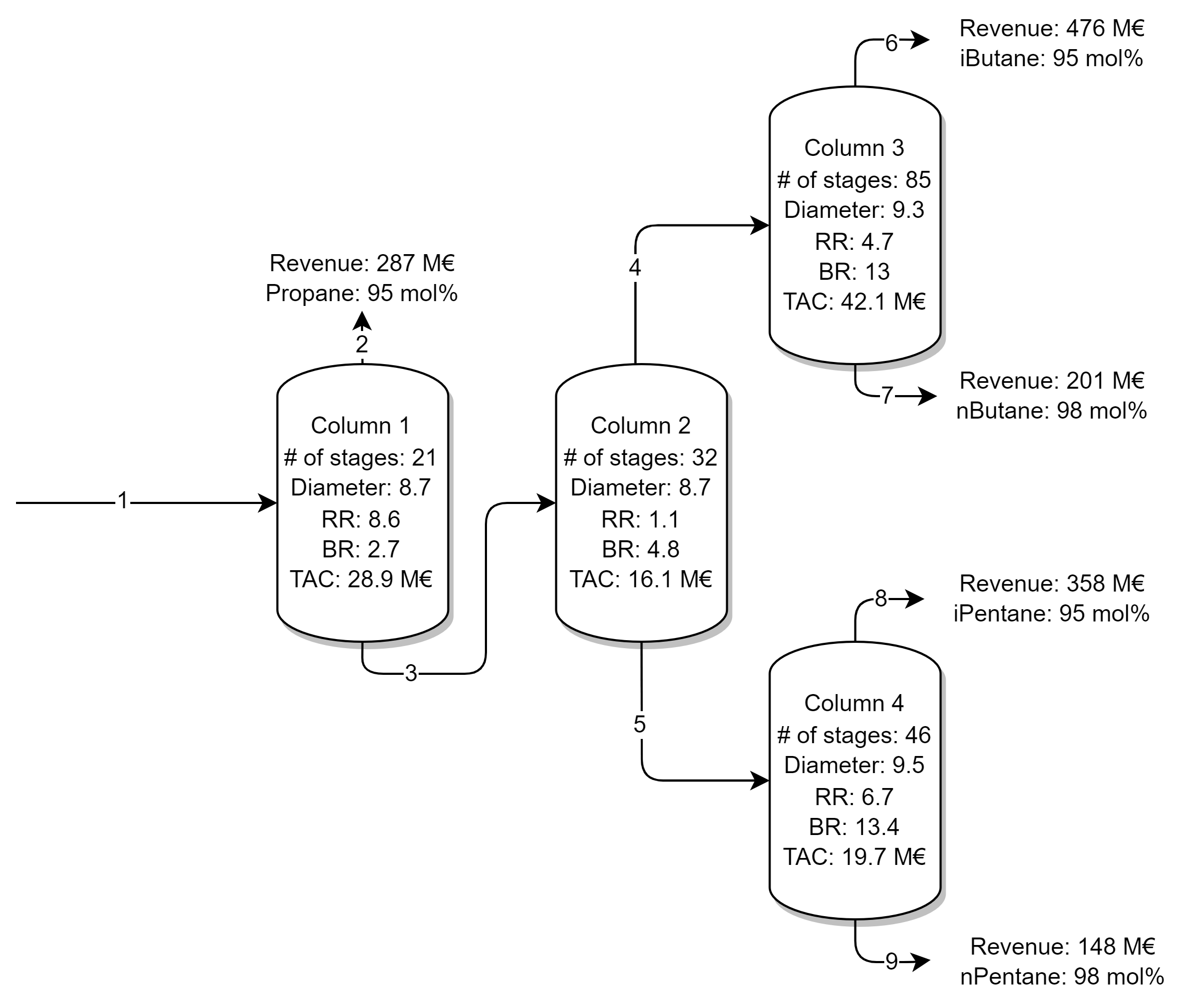}
    \caption{Base case tree-like separation sequence}
    \label{fig:base_case_tree}
\end{figure}

\begin{table}[!htbp]
\centering
\caption{Tree-like column specifications with optimised reflux ratio and distillate/feed ratio}
\label{tree_column_spec}
\begin{tabular}{lccccc}
\multicolumn{1}{c}{Column} & Unit      & 1    & 2    & 3    & 4    \\
Condenser pressure         & {[}bar{]} & 17.2 & 7.2  & 6.7  & 2.0  \\
Reflux ratio     & {[}-{]}   & 8.6  & 1.1  & 4.7  & 6.7  \\
Boilup ratio     & {[}-{]}   & 2.7  & 4.8  & 13.3 & 13.4 \\
Distillate/Feed  & {[}-{]}   & 0.34 & 0.77 & 0.71 & 0.66 \\
Condenser duty   & {[}MW{]}  & -139 & -64  & -126 & -61  \\
Reboiler duty    & {[}MW{]}  & 87   & 49   & 125  & 56  
\end{tabular}
\end{table}

\begin{table}
\centering
\caption{Stream table tree-like base case}
\begin{tabular}{lccllllllllll} 
\toprule
\multicolumn{13}{c}{\begin{tabular}[c]{@{}c@{}}\textbf{Molar purity} \\\textbf{[mol\%]}\end{tabular}} \\ 
\hline
\multicolumn{1}{c}{Stream number} & 1  & 2  & \multicolumn{2}{c}{3}  & \multicolumn{2}{c}{4} & \multicolumn{2}{c}{5 }  & 6    & 7    & 8    & 9     \\
C$_2$H$_6$  & 0.1   & 0.1   & \multicolumn{2}{c}{0.0} & \multicolumn{2}{c}{0.0} & \multicolumn{2}{c}{0.0 }        & 0.0  & 0.0  & 0.0  & 0.0   \\
C$_3$H$_8$  & 33.2  & 95.0  & \multicolumn{2}{c}{1.2}  & \multicolumn{2}{c}{1.6} & \multicolumn{2}{c}{0.0 }        & 2.3  & 0.0  & 0.0  & 0.0   \\
iC$_4$H$_{10}$        & 35.8  & 4.6   & \multicolumn{2}{c}{52.0} & \multicolumn{2}{c}{67.3}        & \multicolumn{2}{c}{0.1 }        & 95.0 & 0.7  & 0.2  & 0.0   \\
nC$_4$H$_{10}$        & 15.4  & 0.3   & \multicolumn{2}{c}{23.3} & \multicolumn{2}{c}{29.8}        & \multicolumn{2}{c}{1.0 }        & 2.7  & 95.0 & 1.5  & 0.0   \\
iC$_5$H$_{12}$        & 10.3  & 0.0   & \multicolumn{2}{c}{15.6} & \multicolumn{2}{c}{1.2} & \multicolumn{2}{c}{64.5 }       & 0.0  & 4.1  & 95.0 & 5.0   \\
nC$_5$H$_{12}$        & 5.2   & 0.0   & \multicolumn{2}{c}{7.9}  & \multicolumn{2}{c}{0.1} & \multicolumn{2}{c}{34.3 }       & 0.0  & 0.2  & 3.3  & 95.0  \\ 
\hline
\multicolumn{13}{c}{\begin{tabular}[c]{@{}c@{}}\textbf{Molar flow } \\\textbf{[kmol/h]}\end{tabular}}   \\ 
\hline
Stream number & 1  & 2  & \multicolumn{2}{c}{3}  & \multicolumn{2}{c}{4} & \multicolumn{2}{c}{5 }  & \multicolumn{1}{c}{6}    & \multicolumn{1}{c}{7}    & \multicolumn{1}{c}{8}    & 9     \\
C$_2$H$_6$  & 6  & 6  & \multicolumn{2}{c}{0}  & \multicolumn{2}{c}{0} & \multicolumn{2}{c}{0 }  & \multicolumn{1}{c}{0}    & \multicolumn{1}{c}{0}    & \multicolumn{1}{c}{0}    & 0     \\
C$_3$H$_8$  & 3995  & 3897  & \multicolumn{2}{c}{99} & \multicolumn{2}{c}{99}  & \multicolumn{2}{c}{0 }  & \multicolumn{1}{c}{99}   & \multicolumn{1}{c}{0}    & \multicolumn{1}{c}{0}    & 0     \\
iC$_4$H$_{10}$        & 4312  & 189   & \multicolumn{2}{c}{4123} & \multicolumn{2}{c}{4121}        & \multicolumn{2}{c}{2 }  & \multicolumn{1}{c}{4107} & \multicolumn{1}{c}{13}   & \multicolumn{1}{c}{2}    & 0     \\
nC$_4$H$_{10}$        & 1857  & 12 & \multicolumn{2}{c}{1845} & \multicolumn{2}{c}{1826}        & \multicolumn{2}{c}{18 } & \multicolumn{1}{c}{118}  & \multicolumn{1}{c}{1709} & \multicolumn{1}{c}{18}   & 0     \\
iC$_5$H$_{12}$        & 1240  & 0  & \multicolumn{2}{c}{1239} & \multicolumn{2}{c}{73}  & \multicolumn{2}{c}{1166 }       & \multicolumn{1}{c}{0}    & \multicolumn{1}{c}{73}   & \multicolumn{1}{c}{1136} & 31    \\
nC$_5$H$_{12}$        & 623   & 0  & \multicolumn{2}{c}{623}  & \multicolumn{2}{c}{3} & \multicolumn{2}{c}{620 }        & \multicolumn{1}{c}{0}    & \multicolumn{1}{c}{3}    & \multicolumn{1}{c}{39}   & 581   \\
Total & 12033 & 4104  & \multicolumn{2}{c}{7929} & \multicolumn{2}{c}{6122}        & \multicolumn{2}{c}{1807 }       & \multicolumn{1}{c}{4324} & \multicolumn{1}{c}{1799} & \multicolumn{1}{c}{1195} & 611   \\ 
\hline
Temperature [°C]        & \multicolumn{1}{l}{105}   & \multicolumn{1}{l}{51.6}  & 105   & 65.0 & 55.3 & 51.3 & 102  & 51.8 & 48.5 & 62.9 & 48.4 & 57.8  \\
Pressure [bar]  & \multicolumn{1}{l} {17.2} & \multicolumn{1}{l}{ 17.2 } &  17.2  & 7.2 & 7.2 & 6.7 & 7.2 & 2.0 & 6.7 & 6.7 & 2.0 & 2.0  \\
\bottomrule
\end{tabular}
\end{table}

\section{Reinforcement learning agent results - max profit}
\label{app:max_profit}
\begin{figure}[H]
    \centering
    \includegraphics[angle=90,origin=c]{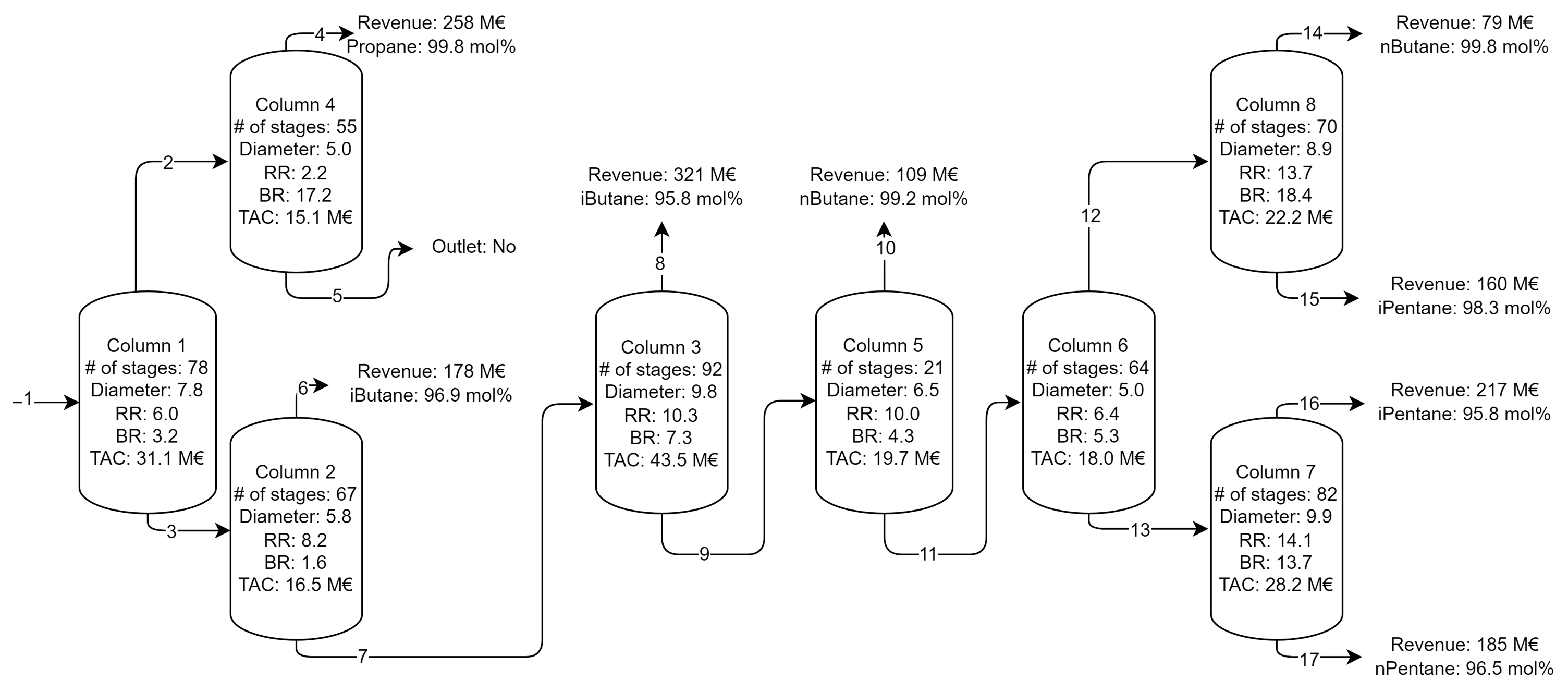}
    \caption{Reinforcement learning separation sequence - max profit}
    \label{fig:max_profit}
\end{figure}

\begin{figure}[H]
    \centering
    \includegraphics[scale=0.6]{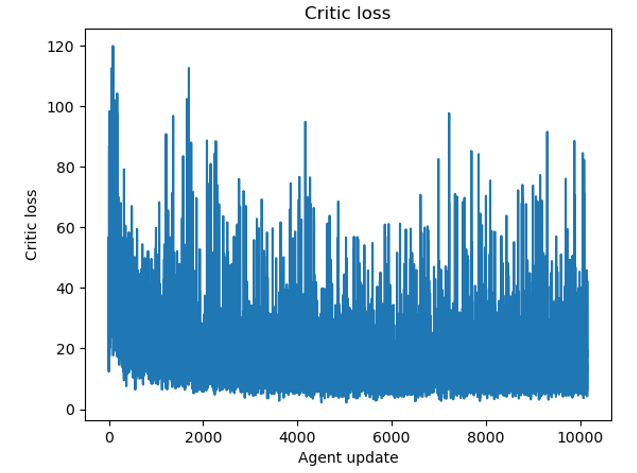}
    \caption{Critic loss - max profit}
    \label{fig:critic_loss_max_profit}
\end{figure}

\begin{figure}[H]
    \centering
    \includegraphics[scale=0.6]{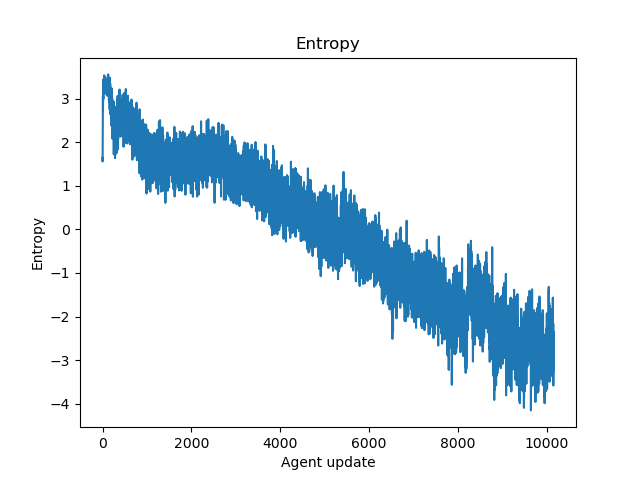}
    \caption{Entropy - max profit}
    \label{fig:entropy_max_profit}
\end{figure}
\newpage
\section{Reinforcement learning agent results - best average}
\label{app:best average}
\begin{figure}[!ht]
    \centering
    \includegraphics[scale=0.6]{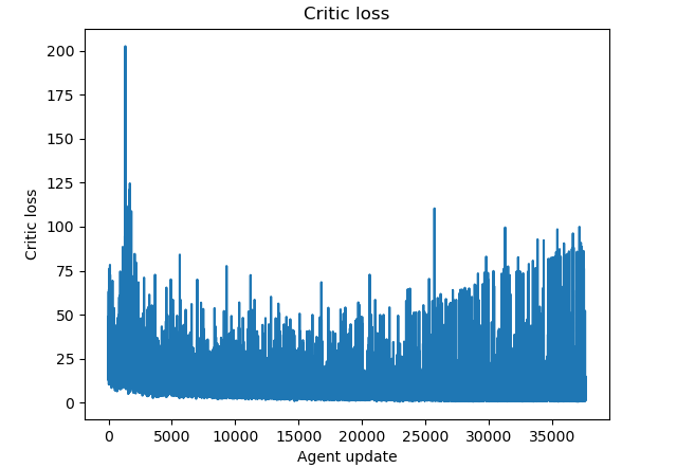}
    \caption{Critic loss - Best average}
    \label{fig:critic loss best average}
\end{figure}

\begin{figure}[H]
    \centering
    \includegraphics[scale=0.6]{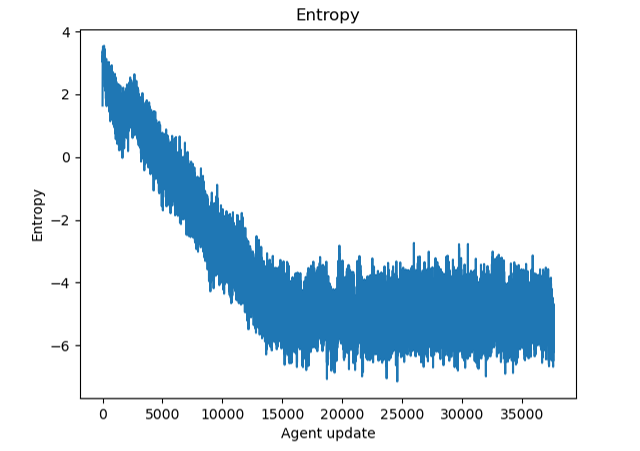}
    \caption{Entropy - best average}
    \label{fig:entropy_best_average}
\end{figure}

\end{appendices}

\end{document}